\documentclass[runningheads]{llncs}
\usepackage{graphicx}
\usepackage{cite}
\usepackage{textcomp}
\usepackage{xcolor}
\usepackage{algorithm}
\usepackage[noend]{algpseudocode}
\algrenewcommand{\algorithmiccomment}[1]{\hskip1em$\#$ #1}
\usepackage[colorlinks = true,linkcolor = blue, urlcolor  = blue, citecolor = blue, anchorcolor = blue]{hyperref}
\usepackage{url}
\usepackage{multirow}
\usepackage{xcolor}
 
\begin{document}

\title{Coevolving Artistic Images Using OMNIREP\thanks{This work was supported by National Institutes of Health (USA) grants LM010098, LM012601, AI116794.
\textcopyright Springer Nature Switzerland AG 2020
J. Romero et al. (Eds.): EvoMUSART 2020, LNCS 12103, pp. 165–178, 2020.
\protect\url{https://doi.org/10.1007/978-3-030-43859-3_12}
}}

\author{Moshe Sipper\inst{1,2} \and
Jason H. Moore\inst{1} \and
Ryan J. Urbanowicz\inst{1}}
\authorrunning{M. Sipper et al.}

%\author{\,}

\institute{\scriptsize Institute for Biomedical Informatics, University of Pennsylvania, Philadelphia, PA 19104, USA \and
Department of Computer Science, Ben-Gurion University, Beer Sheva 84105, Israel \\
\email{sipper@gmail.com},
\url{https://epistasis.org/} }

\maketitle

\begin{abstract}
We have recently developed OMNIREP, a coevolutionary algorithm to discover \textit{both} a representation and an interpreter that solve a particular problem of interest.
Herein, we demonstrate that the OMNIREP framework can be successfully applied within the field of evolutionary art. Specifically, we coevolve representations that encode image position, alongside interpreters that transform these positions into one of three pre-defined shapes (chunks, polygons, or circles) of varying size, shape, and color. We showcase a sampling of the unique image variations produced by this approach. 


\end{abstract}

\section{Introduction}
\label{sec:intro}

Evolutionary art is a branch of evolutionary computation (EC) wherein artwork is generated through an evolutionary algorithm. It is a growing domain, which has boasted a specialized conference over the past few years \cite{Correia:2017} and many 
impressive results \cite{Romero2008,dipaola2009incorporating,Romero2016,wiki:Evolutionary_art}. 

In the present study we focus on the evolution of artistic images. To this end, there are generally three major branches of artistic image evolution, differentiated by the standard of `beauty' applied to drive the fitness of evolving images. The first relies on subjective, interactive feedback from a user \cite{dawkins1996blind,sims1991artificial,secretan2008picbreeder}. 

The second approach relies on a target `inspiration' image to drive fitness. For example, work by \cite{Johansson:2008} used what is essentially a $1+1$ evolution strategy---single parent, single child, both competing against each other---to evolve a replica of the Mona Lisa using semi-transparent polygons. This type of evolved art can produce beautiful abstractions of existing pieces of art, or potentially hybrids of existing images. A series of images sampled during the evolutionary process targeting an inspiration image can also offer an artistically appealing output.

The third approach to  artistic  image  evolution incorporates aesthetic measures in the fitness function (light, saturation, hue, symmetry, complexity, entropy, and more) \cite{greenfield2002simulated,machado2008experiments,den2010comparing,datta2006studying}.

Returning to the evolutionary methodology, one of the EC practitioner's foremost tasks is to identify a representation---a data structure---and its interpretation, or encoding. These can be viewed, in fact, as two distinct tasks, though they are usually dealt with simultaneously. To wit, one might define the representation as a bitstring and in the same breath go on to state the encoding or interpretation, e.g., ``the 120-bit bitstring represents 6 numerical values, each encoded by 20 bits, which are treated as signed floating-point values''.

We have recently developed OMNIREP, a coevolutionary algorithm framework to discover \textit{both} a representation and an interpreter that solve a particular problem of interest \cite{Sipper:OMNIREP}. We applied OMNIREP successfully to regression and program-evolution tasks.
Herein, we demonstrate that OMNIREP can be fruitfully applied within the field of evolutionary art. While the interpreter-representation distinction is perhaps less striking here than with other problems studied by us, we believe both the results and the future possibilities are worthy of presentation, as they demonstrate the efficacy of an alternative and flexible framework for generating evolved art. To the best of our knowledge, this is the first evolutionary art strategy adopting a cooperative (mutualistic) coevolutionary approach, however, some previous work has explored the application of competitive (host-parasite) coevolution to evolving images \cite{greenfield2002simulated}.

In the next section we discuss coevolution and its use as a basis for OMNIREP. 
Section~\ref{sec:prev} briefly discusses some previous work.
Section~\ref{sec:omnirep} presents the OMNIREP algorithm and its application to  evolving artful pictures. 
Results are shown in Section~\ref{sec:results}, followed by
concluding remarks in Section~\ref{sec:conc}.

\section{Coevolution and OMNIREP}
In this paper we consider two tasks---discovering a representation and discovering an interpreter---as distinct yet tightly coupled: \textit{A representation is meaningless without an interpretation; an interpretation is useless without a representation.}
Our basic idea herein is to employ coevolution to evolve the two simultaneously.

\begin{figure}
\centering
\begin{tabular}{ccc}
\includegraphics[height=0.21\textwidth]{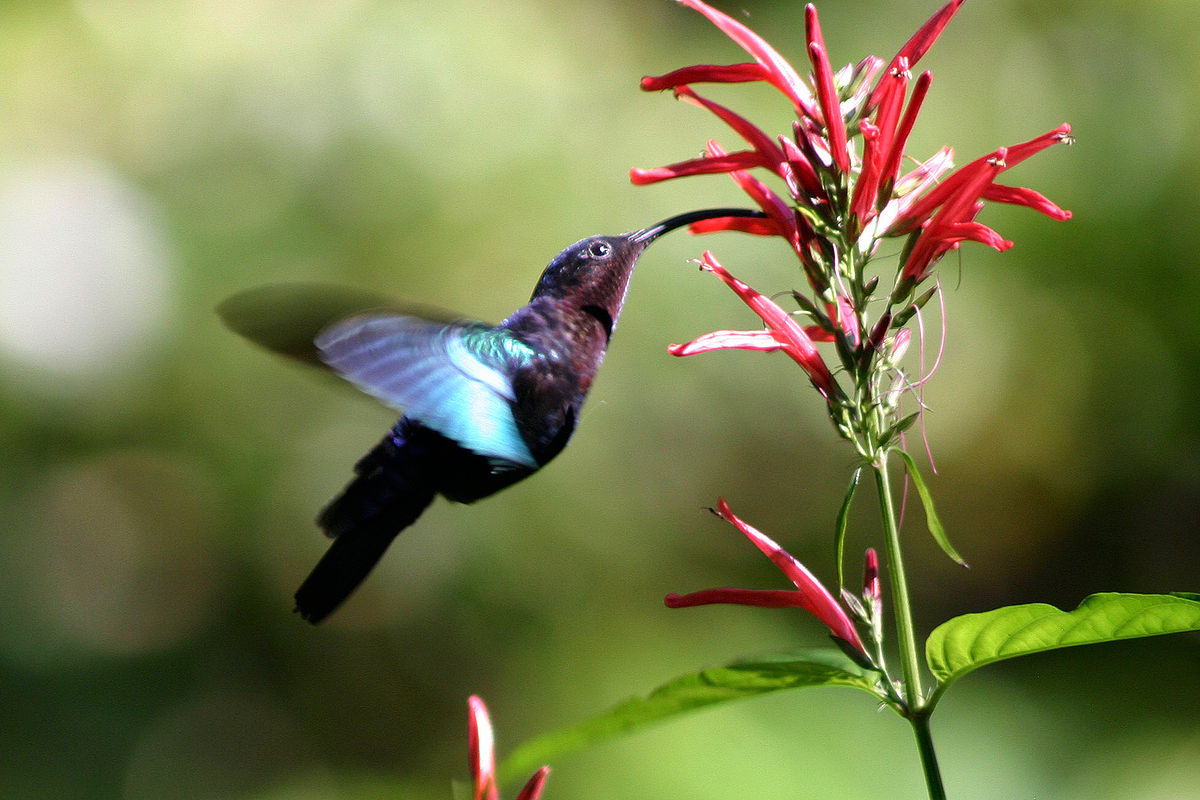} &
\includegraphics[height=0.21\textwidth]{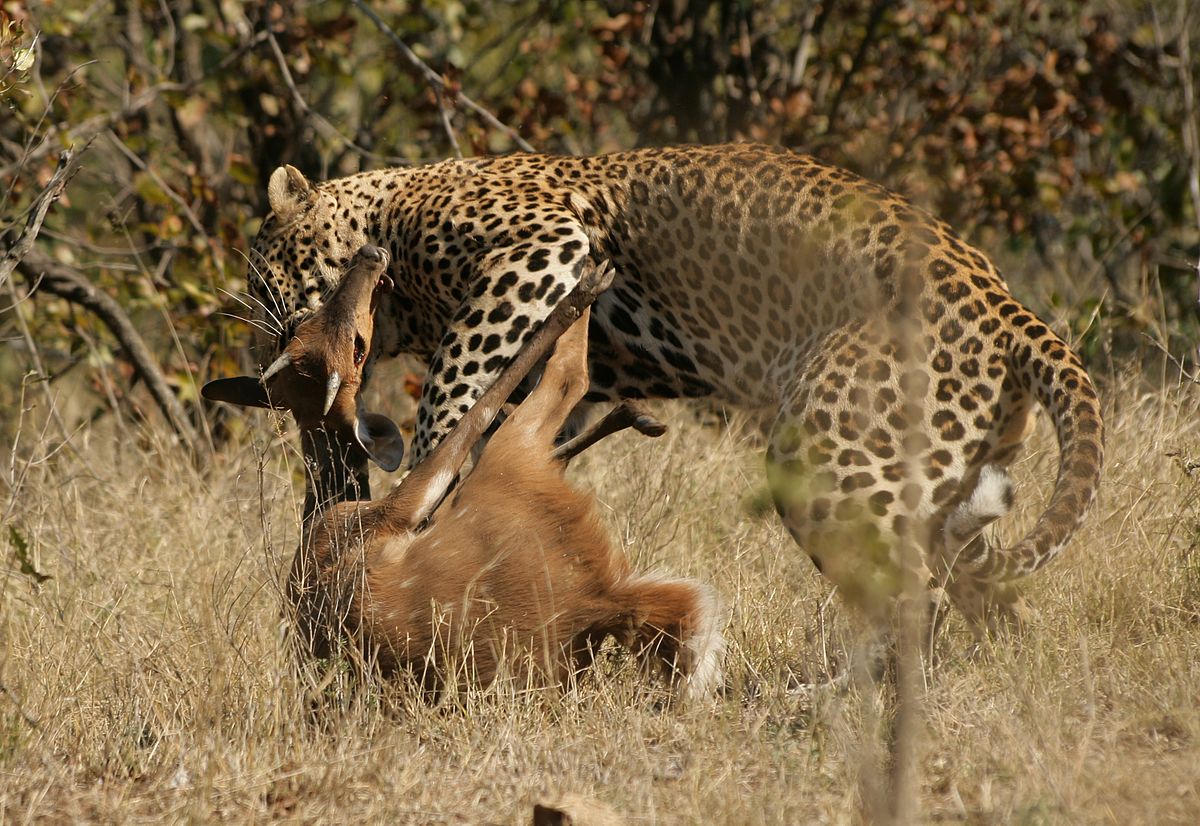} &
\includegraphics[height=0.21\textwidth]{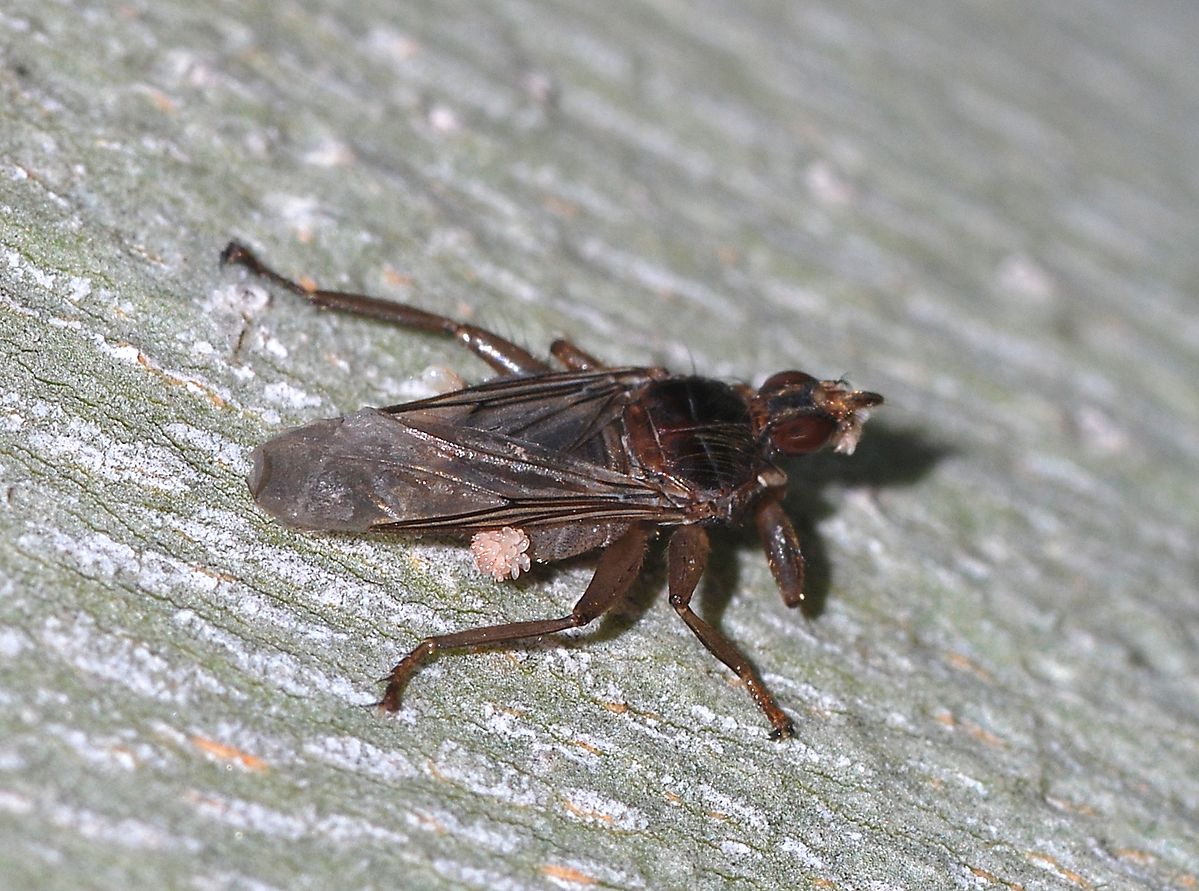} \\
(a) & (b) & (c) \\
\end{tabular}
\caption{Coevolution: 
(a) cooperative: Purple-throated carib feeding from and pollinating a flower (credit: Charles J Sharp,
\protect\url{https://commons.wikimedia.org/wiki/File:Purple-throated_carib_hummingbird_feeding.jpg});
(b) competitive: predator and prey---a leopard killing a bushbuck
(credit: NJR ZA, \protect\url{https://commons.wikimedia.org/wiki/File:Leopard_kill_-_KNP_-_001.jpg});
(c) commensalistic: Phoretic mites attach themselves to a fly for transport
(credit: Alvesgaspar, \protect\url{https://en.wikipedia.org/wiki/File:Fly_June_2008-2.jpg}).
} \label{fig:coevolution}
\end{figure}

Coevolution refers to the simultaneous evolution of two or more species with coupled fitness \cite{Pena:2001}. Coevolving species usually compete or cooperate, with a third form of coevolution being commensalistism, wherein members of one species gain benefits while those of the other species neither benefit nor are harmed \cite{Sipper:SAFE} (Figure~\ref{fig:coevolution}). 

In a competitive coevolutionary algorithm the fitness of an individual is based on direct competition with individuals of other species, which in turn evolve separately in their own populations. Increased fitness of one of the species implies a reduction in the fitness of the other species \cite{Hillis:1990}. 

A cooperative coevolutionary algorithm involves a number of independently evolving species, which come together to obtain problem solutions. The fitness of an individual depends on its ability to collaborate with individuals from other species \cite{Pena:2001,Potter:2000,Dick:2014}. 

The basic idea of OMNIREP can be stated simply: Rather than specify a specific representation along with a specific interpreter in advance, we shall set up a cooperative coevolutionary algorithm to coevolve the two, with a population of representations coevolving alongside a population of interpreters.\footnote{OMNIREP derives from `OMNI'---universal, and `REP'---representation; it also denotes an acronym: Originating MeaNing by coevolving Interpreters and REPresentations.}

Of importance to note is OMNIREP's not being a specific algorithm but rather an algorithmic framework, which can hopefully be of use in various settings. We believe that the OMNIREP methodology can aid researchers not only in solving \textit{specific} problems but also as an \textit{exploratory} tool when one is seeking out a good representation \cite{Sipper:OMNIREP}. 

\section{Previous Work}
\label{sec:prev}
Generative and Developmental Encoding is a branch of EC concerned with genetic encodings motivated by biology. A structure that repeats multiple times can be represented by a single set of genes that is reused in a genotype-to-phenotype mapping \cite{stanley:2003,Hart:1995,Gruau:1996,Bentley:1999,Koza:2003,Koza:1999,Hornby:2002,Stanley:2009}.

In Gene Expression Programming the individuals in the population are encoded as linear strings of fixed length, which are afterwards expressed as nonlinear entities of different sizes and shapes (i.e., simple diagram representations or expression trees)
\cite{Ferreira:2001}.

Though not used extensively, variable-length genomes have been around for quite some time (of course, some representations, such as trees in genetic programming, are inherently variable-length; herein, we simply refer to the literature on ``variable-length genomes'') \cite{Goldberg:1989m,Lee:2000}.

Grammatical Evolution (GE) was introduced by \cite{Ryan:1998} as a variation on genetic programming. Here, a Backus-Naur Form (BNF) grammar is specified that allows a computer program or model to be constructed by a simple genetic algorithm operating on an array of bits. The GE approach is appealing because only the specification of the grammar needs to be altered for different applications. One might consider subjecting the grammar encoding to evolution in an OMNIREP manner (as done, e.g., by \cite{Azad:2006}). 

Within a memetic computing framework, Iacca et al. \cite{iacca2012ockham} proposed, ``a bottom-up approach which starts constructing the algorithm from scratch and, most importantly, allows an understanding of functioning and potentials of each search operator composing the algorithm.'' 
Caraffini et al. \cite{caraffini2014analysis} proposed a computational prototype for the automatic design of optimization algorithms, consisting of two phases: a problem analyzer first detects the features of the problem, which are then used to select the operators and their links, thus performing the algorithmic design automatically. Both these works share the desire to tackle basic algorithmic design issues in a (more) automatic manner.

Tangentially related to our work herein is the extensive research on parameters and hyper-parameters in EC, some of which has focused on self-adaptive algorithms, wherein the parameters to be adapted are encoded into the chromosomes and undergo crossover and mutation. The reader is referred to \cite{Sipper:2018} for a comprehensive discussion of this area.
Another tangential connection is to ``smart'' crossover and mutation operators, wherein, interestingly, coevolution has also been applied \cite{Zaritsky:2004}.

\section{Evolving Art using OMNIREP}
\label{sec:omnirep}

\begin{figure}
    \centering
    \includegraphics[width=1.0\textwidth]{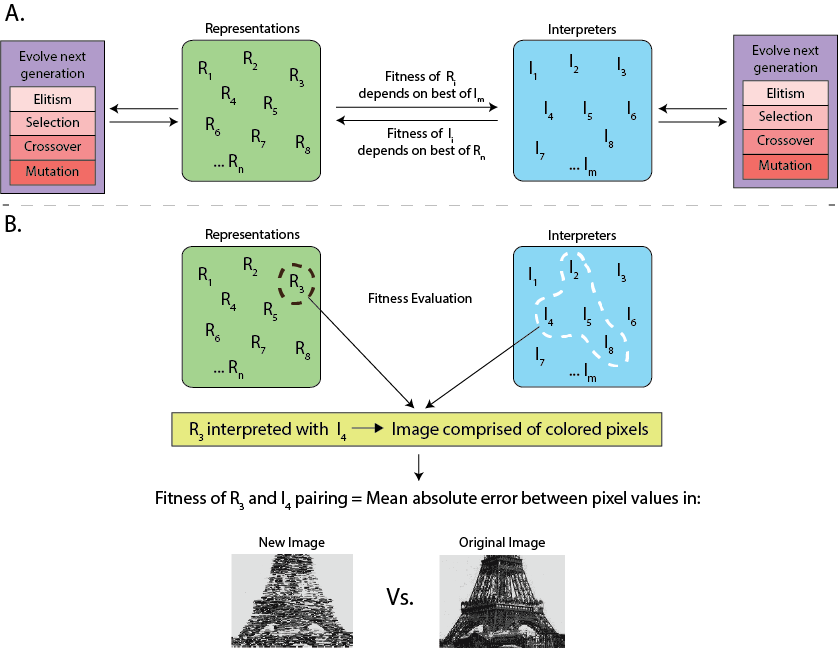}
    \caption{OMNIREP algorithm adapted to the task of evolutionary art. (A). OMNIREP includes two coevolving populations, one with candidate representations, and the other with candidate interpreters. Each population is evolved using the same fundamental evolutionary algorithm mechanisms (as summarized in the purple boxes). (B) The fitness of a given representation depends on representative interpreters (i.e., the 4 interpreters with the best fitness from the previous generation). In this example $R_{3}$'s fitness is the average fitness of the four representation-interpreter pairings. The fitness of a given interpreter (e.g., $I_{3}$) similarly depends on representative representations (not shown). 
    The fitness of a representation-interpreter pair is computed by combining a representation individual (R) with an interpreter individual (I) to produce the pixel values of an image. Pair fitness is the mean absolute error between the pixels of the new image vs. the  inspiration image.}
    \label{fig:omnirep-e}
\end{figure}

OMNIREP uses cooperative coevolution with two coevolving populations, one of representations, the other of interpreters. The evolution of each population is identical to a single-population evolutionary algorithm---except where fitness is concerned (Figure~\ref{fig:omnirep-e}).
One might argue that the distinction between ``representation'' and ``interpretation'' is a malleable one, but this distinction is often put to good use in computer science \cite{Sipper:OMNIREP}.

We describe below in detail the components and parameters of the OMNIREP system, including:
population composition, initialization, selection, crossover, mutation, fitness, elitism, evolutionary rates, and parameters.

\begin{figure}
    \centering
    \includegraphics[width=0.9\textwidth]{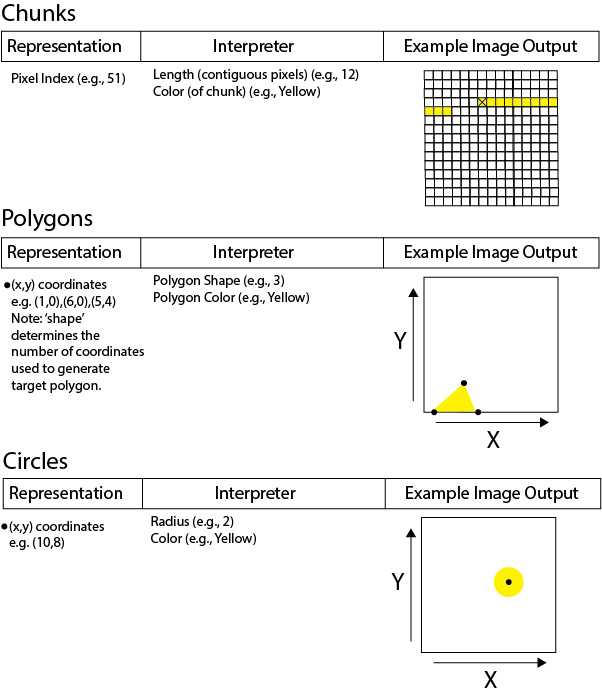}
    \caption{Overview and examples of the three image-mapping strategies employed by OMNIREP in experimental evolutionary runs. }
    \label{fig:omnirep-shape}
\end{figure}

\textbf{Populations.}
To showcase the application of OMNIREP to evolutionary art, we designed three relatively simple interpreter-representation setups, which produced quite striking results; we refer to them as: chunks, polygons, and circles (Figure~\ref{fig:omnirep-shape}). These setups evidence the ease with which OMNIREP can be applied beneficially.

\textit{Chunks.}
An evolving image is composed of linear chunks.
A two-dimensional image of dimensions $\{width,height\}$ is treated as a one-dimensional list of pixels of size $width \times height$ (ranging from 10848 pixels to 68816, depending on the particular inspiration image selected for fitness). The representation individual's genome is a list of pixel indexes, $p_i \in \{0,\ldots,width*height-1\}$, $i= 1,\ldots,5000$, where $p_i$ is the start of a same-color chunk of pixels. 
The interpreter individual is a list equal in length to the representation individual, consisting of tuples $(b_i,c_i)$, where $b_i$ is chunk $i$'s length, and $c_i$ is chunk $i$'s color. 
The image-producing process moves sequentially through the list of chunks, coloring pixels $i$ through $i+b_i-1$ with color $c_i$.
If a pixel is uncolored by any chunk it is assigned a default base color.

Thus an interpreter individual combines with a representation individual to paint a picture, made up of same-color chunks of length and color indicated by the former and start positions indicated by the latter.

\textit{Polygons.}
An evolving image is composed of polygons.
The representation individual's genome is a list of 600 polygon coordinates, 
$[x_i,y_i]$, $i \in \{1,\ldots,600\}$.
The number of polygons was set to 50 and the maximum number of sides to 12, hence 600.
The interpreter individual is a list of length 50, representing 50 polygons, consisting of tuples $(s_i,c_i)$, where $s_i$ is polygon $i$'s shape, i.e., number of sides, and $c_i$ is polygon $i$'s color.

An interpreter individual combines with a representation individual to paint a picture, made up of 50 polygons whose shape and color  are indicated by the former and coordinates by the latter. 
The image-producing process moves sequentially through the interpreter and representation genomes, picking the appropriate number of coordinates determined by $s_i$, and coloring the resultant shape with color $c_i$. Coordinates ``left over'' in the representation genome in the end are unused.

\textit{Circles.}
An evolving image is composed of circles.
The representation individual's genome is a list of 50 circle centers, 
$[x_i,y_i]$, $i \in \{1,\ldots,50\}$.
The interpreter individual is a list equal in length to the representation individual, consisting of tuples $(c_i,r_i)$, where $c_i$ is circle $i$'s color and $r_i$ is its radius.

An interpreter individual combines with a representation individual to paint a picture, made up of 50 circles whose colors and radii are indicated by the former and centers by the latter.

For chunks, polygons, and circles, if there is any overlap in these shapes forming the image, the most recently placed shape opaquely covers the overlapped shape. 

\begin{figure}
    \centering
    \begin{tabular}{cccc}
      \includegraphics[height=0.17\textwidth]{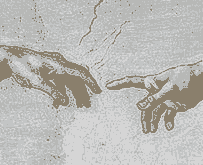} &  \includegraphics[height=0.17\textwidth]{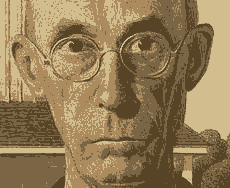} &
      \includegraphics[height=0.17\textwidth]{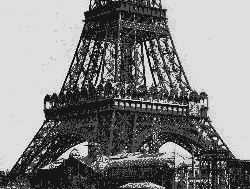} &  \includegraphics[height=0.17\textwidth]{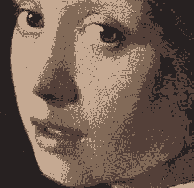} \\
      & & & \\
      \includegraphics[height=0.17\textwidth]{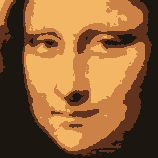} & 
      \includegraphics[height=0.17\textwidth]{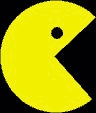} &
        \includegraphics[height=0.17\textwidth]{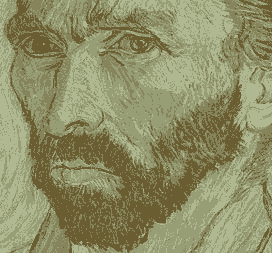} & 
        \includegraphics[height=0.17\textwidth]{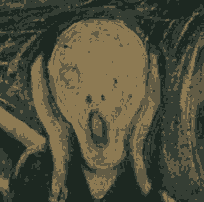} \\
   \end{tabular}
    \caption{Inspirational images.}
    \label{fig:orig}
\end{figure}

\textbf{Initialization.}
For every coevolutionary run: both populations are initialized to random values in the appropriate range, depending on the particulars of the interpreter-representation setup delineated above (chunks, polygons, or circles). 
An inspiration image is chosen per evolutionary run as one of the 8 shown in Figure~\ref{fig:orig}. The inspiration images were converted to 4 colors from their originals using Python's \texttt{Image.ADAPTIVE} palette (in the \texttt{PIL} package). Note that the 4 colors differ between images.

\textbf{Selection.} 
Tournament selection with tournament size 4, i.e., choose 4 individuals at random from the population and return the individual with the best fitness as the selected one.

\textbf{Crossover.} Single-point crossover --- select a random crossover point and swap two parent genomes beyond this point to create two offspring --- is employed every generation. 

\textbf{Mutation.}
Mutation is done with probability 0.3 (per individual) by selecting a random gene and replacing it with a new random value in the appropriate range.

\textbf{Fitness.}
To compute fitness the two coevolving populations cooperate. Specifically, to compute the fitness of a single individual in one population, we use \textit{representatives} from the other population \cite{Pena:2001}. The representatives (also called cooperators) are selected via a greedy strategy as the 4 fittest individuals from the previous generation. When evaluating the fitness of a particular representation individual, we combine it 4 times with the top 4 interpreter individuals, compute 4 fitness values, and use the average fitness over these 4 evaluations as the final fitness value of the representation individual. In a similar manner we use the average of 4 representatives from the representations population when computing the fitness of an interpreter individual. (Other possibilities include using the best fitness of the 4, the worst fitness, and selecting a different mix of representatives, e.g.,  best, median, and worst.)

A representation individual and an interpreter individual are combined as described above per interpreter-representation  setup  (chunks, polygons, or circles). A single fitness value then equals the mean absolute error with respect to the known inspirational pixels (Figure~\ref{fig:omnirep-e}).

Note that our objective was not to reproduce the original image precisely---which would be uninteresting. Rather, the selected image serves as inspiration, setting an evolutionary direction.

\textbf{Elitism.} The 2 individuals with the highest fitness in a generation are copied (``cloned'') into the next generation unchanged.

\textbf{Evolutionary rates} differ between the two populations, with the interpreters population evolving more slowly, specifically, every 3 generations.

\textbf{Parameters.}
Table~\ref{tab:params} provides a summary of parameters discussed throughout the paper.\footnote{Some parameters may seem arbitrary but our recent findings provide some justification for this \cite{Sipper:2018}.}

\begin{table}
    \centering
    \caption{Evolutionary parameters. Shown first are common parameters, followed by experiment-specific ones.}
    \footnotesize
    \begin{tabular}{r@{\hskip 20px}l}
      Description & Value \\ 
      \hline 
      \multicolumn{1}{l}{Common} & \\
%         Number of evolutionary runs per image & 10  \\
         Number of images & 8  \\
         Size of representations population & 20 \\
         Size of interpreters population & 10 \\
         Type of selection & Tournament \\
         Tournament size & 4 \\
         Type of crossover & single-point \\
         Probability of mutation (representations) & 0.3 \\
         Probability of mutation (interpreters) & 0.3 \\
         Evolve interpreters population every & 3 generations \\
         Number of representatives used for fitness & 4 \\
         Number of top individuals copied (elitism) & 2 \\ 
         Number of colors & 4 \\ 
         \hline
      \multicolumn{1}{l}{Chunks} & \\
         Number of generations & 20000 \\
         Size of representation individual & 5000 \\ 
         Size of interpreter individual & 5000 (chunks) \\
         Minimum chunk size & 1 pixel \\
         Maximum chunk size & 10 pixels \\ 
         \hline
      \multicolumn{1}{l}{Polygons} & \\ 
         Number of generations & 50000 \\
         Size of representation individual & 600  \\ 
         Size of interpreter individual & 50 (polygons) \\
         Minimum polygon sides & 3 \\
         Maximum polygon sides & 12 \\ 
         \hline
      \multicolumn{1}{l}{Circles} & \\
         Number of generations & 50000 \\
         Size of representation individual & 50 \\ 
         Size of interpreter individual & 50 (circles) \\
         Minimum radius & 3 \\
         Maximum radius & 50 \\
    \end{tabular}
    \normalsize
    \label{tab:params}
\end{table}

% Our goal herein was to evolve artistic renderings of given images. We were inspired by the work of \cite{Johansson:2008}, who used what is essentially a $1+1$ evolution strategy---single parent, single child, both competing against each other---to evolve a replica of the Mona Lisa using semi-transparent polygons. 

\section{Results}
\label{sec:results}
\begin{figure}
    \centering
    \begin{tabular}{cccc}
      \includegraphics[height=0.17\textwidth]{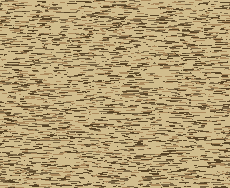} &  \includegraphics[height=0.16\textwidth]{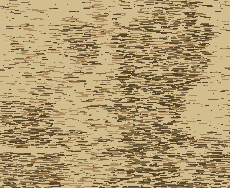} &
      \includegraphics[height=0.17\textwidth]{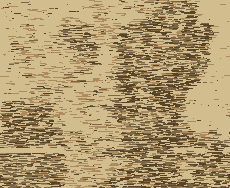} &  \includegraphics[height=0.16\textwidth]{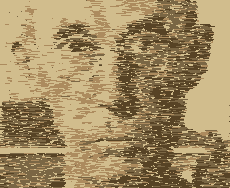} \\
      &  &  &  \\
      \includegraphics[height=0.17\textwidth]{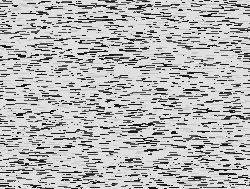} &  \includegraphics[height=0.16\textwidth]{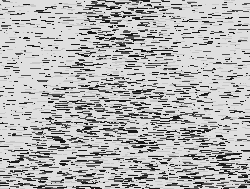} &
      \includegraphics[height=0.17\textwidth]{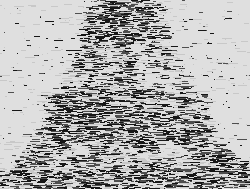} &  \includegraphics[height=0.16\textwidth]{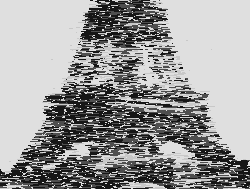} \\
      &  &  &  \\
      \includegraphics[height=0.17\textwidth]{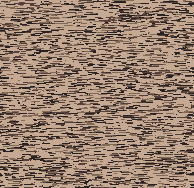} &  \includegraphics[height=0.16\textwidth]{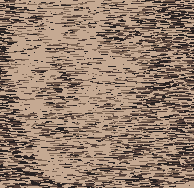} &
      \includegraphics[height=0.17\textwidth]{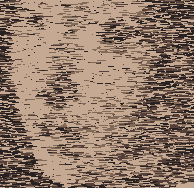} &  \includegraphics[height=0.16\textwidth]{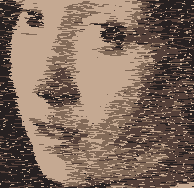} \\
      &  &  &  \\
      \includegraphics[height=0.17\textwidth]{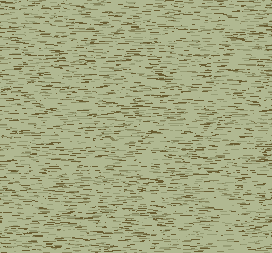} &  
      \includegraphics[height=0.17\textwidth]{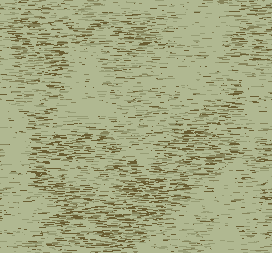} &
      \includegraphics[height=0.17\textwidth]{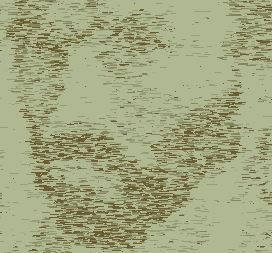} &  
      \includegraphics[height=0.17\textwidth]{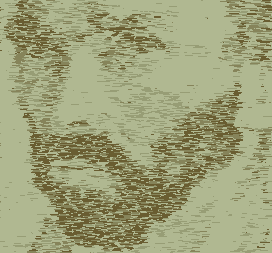} \\
   \end{tabular}
    \caption{Evolutionary trajectories of evolving images (chunks). Each row represents a single run.}
    \label{fig:chunks}
\end{figure}

\begin{figure}
    \centering
    \begin{tabular}{cccc}
      \includegraphics[height=0.17\textwidth]{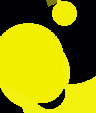} &  
      \includegraphics[height=0.17\textwidth]{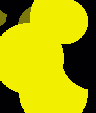} &
      \includegraphics[height=0.17\textwidth]{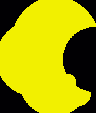} &  
      \includegraphics[height=0.17\textwidth]{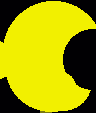} \\
      &  &  &  \\
      \includegraphics[height=0.17\textwidth]{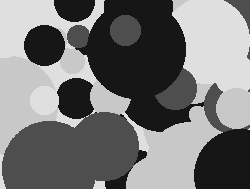} &  
      \includegraphics[height=0.17\textwidth]{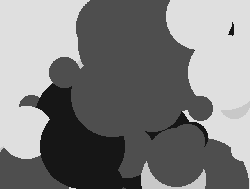} &
      \includegraphics[height=0.17\textwidth]{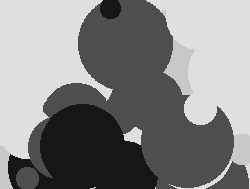} &  
      \includegraphics[height=0.17\textwidth]{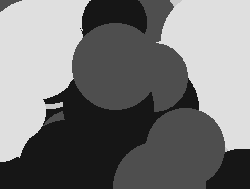} \\
   \end{tabular}
    \caption{Evolutionary trajectories of evolving images (circles). Each row represents a single run.}
    \label{fig:circles}
\end{figure}

\begin{figure}
    \centering
    \begin{tabular}{cccc}
      \includegraphics[height=0.17\textwidth]{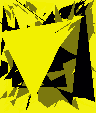} &  
      \includegraphics[height=0.17\textwidth]{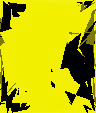} &
      \includegraphics[height=0.17\textwidth]{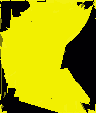} &  
      \includegraphics[height=0.17\textwidth]{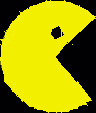} \\
      &  &  &  \\
      \includegraphics[height=0.17\textwidth]{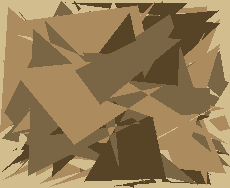} &  
      \includegraphics[height=0.17\textwidth]{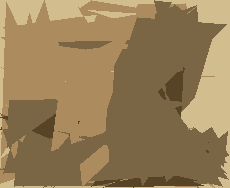} &
      \includegraphics[height=0.17\textwidth]{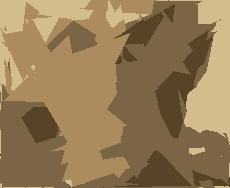} &  
      \includegraphics[height=0.17\textwidth]{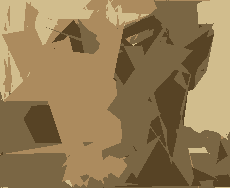} \\
      &  &  &  \\
      \includegraphics[height=0.17\textwidth]{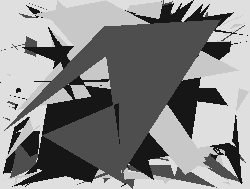} &  
      \includegraphics[height=0.17\textwidth]{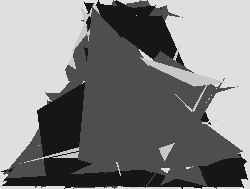} &
      \includegraphics[height=0.17\textwidth]{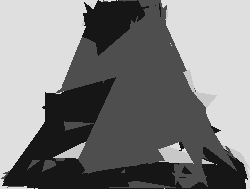} &  
      \includegraphics[height=0.17\textwidth]{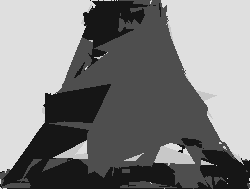} \\
      &  &  &  \\
      \includegraphics[height=0.17\textwidth]{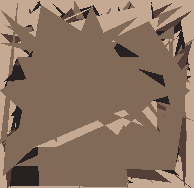} &  
      \includegraphics[height=0.17\textwidth]{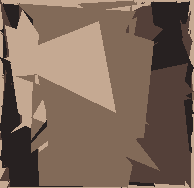} &
      \includegraphics[height=0.17\textwidth]{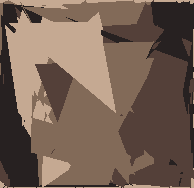} &  
      \includegraphics[height=0.17\textwidth]{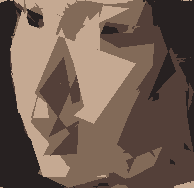} \\
      &  &  &  \\
      \includegraphics[height=0.17\textwidth]{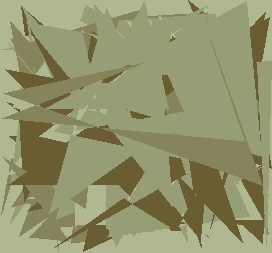} &  
      \includegraphics[height=0.17\textwidth]{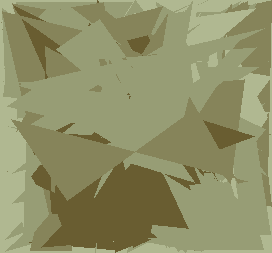} &
      \includegraphics[height=0.17\textwidth]{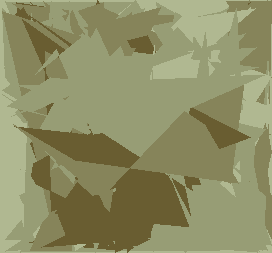} &  
      \includegraphics[height=0.17\textwidth]{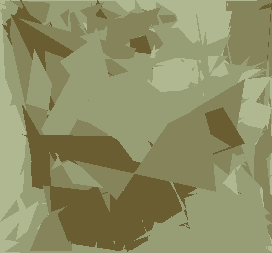} \\
      &  &  &  \\
      \includegraphics[height=0.17\textwidth]{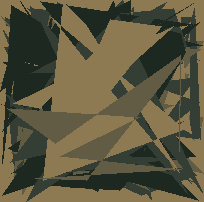} &  
      \includegraphics[height=0.17\textwidth]{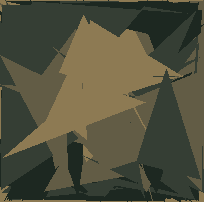} &
      \includegraphics[height=0.17\textwidth]{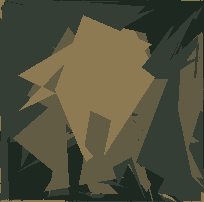} &  
      \includegraphics[height=0.17\textwidth]{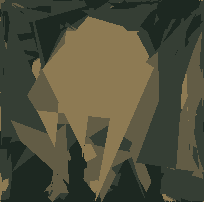} \\
      &  &  &  \\
   \end{tabular}
    \caption{Evolutionary trajectories of evolving images (polygons). Each row represents a single run.}
    \label{fig:polygons}
\end{figure}

We performed multiple evolutionary runs for each of the 8 target images. 
Figures~\ref{fig:chunks} through~\ref{fig:polygons} present select examples of the images evolved with OMNIREP.  Each figure gives sample outputs from each of the three underlying setups utilized (i.e., chunks, circles, or polygons). Each set of 4 images (derived from a specific inspiration image) serves two purposes: (1) demonstrating the evolutionary trajectories of evolving art, namely, images in intermediate generations of an evolutionary run, and (2) presenting the images as part of a `panel', i.e., a collection of images in series, meant to be viewed as a single artistic piece. Indeed, subjectively, we feel that part of the appeal of the images evolved by OMNIREP is the progression of an image from a state of abstraction to one closer to the inspirational piece. Of course, individual images generated with OMNIREP can also be selected as the output artistic piece. 

Beyond generating static images, we have also explored converting the set of evolving images into appealing animated GIFs (see samples at \url{https://github.com/EpistasisLab/OMNIREP}).

\section{Concluding Remarks}
\label{sec:conc}
We adapted and applied OMNIREP, which coevolves representations and interpreters simultaneously, to evolutionary art. This work demonstrates the potential of a mutualistic coevolutionary system, as well as a strategy that separates representations from interpreters, towards the task of evolved artistic images. This framework can be flexibly extended in the future to generate an even greater diversity of novel and intriguing imagery. 

Some of the immediate future directions we expect would be valuable to explore include:
\begin{enumerate}
\item Expansion of the color palette beyond 4 colors per image.

\item Giving interpreters the option to choose one or more shapes (i.e., chunks, polygons, or circles) to incorporate into a given evolved image.

\item Expand the shape options to include other forms or orientations (e.g., ellipsoids, squares, rectangles, etc.).

\item Promote overlapping shapes in the image and assign color mixes to the image in these overlapping regions.

\item More complex interpreters, e.g., an interpreter genetic programming tree that interprets a representation of pixels.

\item Evolving a hybrid of multiple inspiration images through integration of multiobjective optimization.

\item Incorporate novelty \cite{Lehman:2008} into the coevolutionary algorithm (as proposed in \cite{Sipper:SAFE}) to promote a greater variety of novel images that create interesting departures from the inspiration image. 
\end{enumerate}

We perceive OMNIREP not as a particular algorithm but rather as a meta-algorithm, which might hopefully be suitable for other settings, beyond the artful one described herein. Essentially, any scenario where some form of representation may be interpreted in several ways, or where the representation and interpreter can be rendered ``fluid'' rather than fixed, might be a candidate for an OMNIREP approach.

\bibliographystyle{splncs04}
\bibliography{omnirep}
\end{document}